\pdfoutput=1
\documentclass{article}
\usepackage{ismir,amsmath,cite}
\usepackage{graphicx}
\usepackage{color}
\usepackage{subfig}
\usepackage[utf8]{inputenc}

\usepackage{url}

\title{Towards Score Following in Sheet Music Images}

\multauthor
{Matthias Dorfer \hspace{1cm} Andreas Arzt \hspace{1cm} Gerhard Widmer}
{Department of Computational Perception, Johannes Kepler University Linz, Austria\\
{\tt matthias.dorfer@jku.at}
}
\def\authorname{Matthias Dorfer, Andreas Arzt, Gerhard Widmer}

\begin{document}

\maketitle
\begin{abstract}
This paper addresses the matching of short music audio snippets
to the corresponding pixel location in images of sheet music.
A system is presented that simultaneously learns to read notes, listens to music
and matches the currently played music to its corresponding notes in the sheet.
It consists of an end-to-end multi-modal convolutional neural network
that takes as input images of sheet music and spectrograms of the respective audio snippets. It learns to predict, for a given unseen audio snippet
(covering approximately one bar of music),
the corresponding position in the respective score line.
Our results suggest that with the use of (deep) neural networks
-- which have proven to be powerful image processing models --
working with sheet music becomes feasible and a promising future research direction.
\end{abstract}

\section{Introduction}
\label{sec:introduction}
%!TEX root = ISMIR2016template.tex

Precisely linking a performance to its respective sheet music -- commonly
referred to as audio-to-score alignment -- is an important topic in MIR
and the basis for many applications
\cite{Thomas_2012_LinkingAudioAndSheetMusic}. For instance, the
combination of score and audio supports algorithms and
tools that help musicologists in in-depth performance analysis (see e.g. \cite{Cook_2007_Chopin}), allows for new
ways to browse and listen to classical music (e.g. \cite{Dunn_2006_Variations,Melenhorst_2015_tablet}), and can generally be
helpful in the creation of training data for tasks like beat tracking or
chord recognition. When done on-line, the alignment task is known as score
following, and enables a range of applications like the synchronization of
visualisations to the live music during concerts (e.g.
\cite{Arzt_2015_AIConcertgebouw,Prockup_2013_Orchestra}), and automatic
accompaniment and interaction live on stage (e.g.
\cite{Cont_2009_CDFA,Raphael_2010_MusicPlusOne}).

So far all approaches to this task depend on a symbolic, computer-readable
representation of the sheet music, such as MusicXML or MIDI
(see e.g.
\cite{Arzt_2015_AIConcertgebouw,Prockup_2013_Orchestra,Cont_2009_CDFA,Raphael_2010_MusicPlusOne,Mueller_2005_AudioMatching,Niedermayer_2010_PHD,Miron_2014_Alignment,Duan_2011_Alignment,Izmirli_2012_BridPrinMusiAnd}).
This representation is created either manually (e.g. via the time-consuming process
of (re-)setting the score in a music notation program), or automatically via
optical music recognition software. Unfortunately automatic methods are still
highly unreliable and thus of limited use, especially for more complex music like
orchestral scores \cite{Thomas_2012_LinkingAudioAndSheetMusic}.

The central idea of this paper is to develop a method that links the audio and
the image of the sheet music \emph{directly}, by \emph{learning} correspondences
between these two modalities, and thus making the complicated step of creating
an in-between representation obsolete.
We aim for an algorithm that simultaneously learns to \emph{read
notes}, \emph{listens} to music and \emph{matches} the currently played music
with the correct notes in the sheet music. We will tackle the problem in an
end-to-end neural network fashion, meaning that the entire behaviour of the
algorithm is learned purely from data and no further manual feature
engineering is required.

\section{Methods}
\label{sec:methods}
%!TEX root = ISMIR2016template.tex
This section describes the audio-to-sheet matching model and the input data required,
and shows how the model is used at test time to predict the expected location
of a new unseen audio snippets in the respective sheet image.

\subsection{Data, Notation and Task Description}
% We begin by introducing the data and a concise notation to describe our approach.
The model takes two different input modalities at the same time:
images of scores, and short excerpts from spectrograms of audio renditions of the score
(we will call these \emph{query snippets} as the task is to predict the position
in the score that corresponds to such an audio snippet). 
For this first proof-of-concept paper, we make a number of simplifying assumptions:
for the time being, the system is fed only a \emph{single staff line} at a time
(not a full page of score). We restrict ourselves to \emph{monophonic music}, and to the
\emph{piano}.
To generate training examples, we produce a fixed-length query snippet
for each note (onset) in the audio. The snippet covers the target note onset plus a few
additional frames, at the end of the snippet, and a fixed-size context of
$1.2$ seconds into the past, to give some temporal context.
The same procedure is followed when
producing example queries for off-line testing.

\begin{figure}[t!]
\centering
\subfloat[Spectrogram-to-sheet correspondence.
 		  In this example the rightmost onset in spectrogram excerpt $\mathbf{E}_{i,j}$
 		  corresponds to the rightmost note (target note $j$) in sheet image $\mathbf{S}_i$.
 		  For the present case the temporal context of about $1.2$ seconds (into the past)
 		  covers five additional notes in the spectrogram.
 		  The staff image and spectrogram excerpt are exactly the multi-modal input
 		  presented to the proposed audio-to-sheet matching network.
 		  At train time the target pixel location $x_j$ in the sheet image is available;
 		  at test time $\hat{x}_j$ has to be predicted by the model (see figure below).]
 		  {\label{fig:annotation_example}
 		  {\includegraphics[width=0.73\columnwidth]{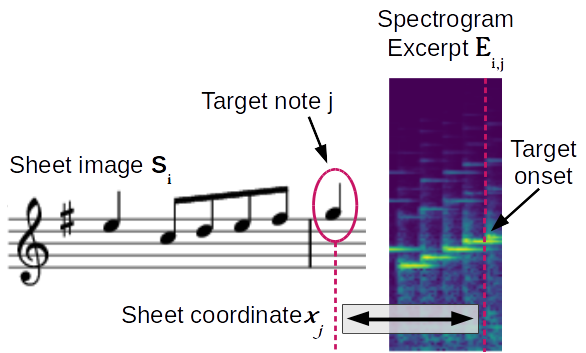}}}
\qquad
\vspace{3px}
\subfloat[Schematic sketch of the audio-to-sheet matching task targeted in this work.
		  Given a sheet image $\mathbf{S}_i$ and a short snippet of audio (spectrogram excerpt $\mathbf{E}_{i,j}$)
		  the model has to predict the audio snippet’s corresponding pixel location $x_j$ in the image.]
		 {\label{fig:task_sketch}
		 {\includegraphics[width=0.9\columnwidth]{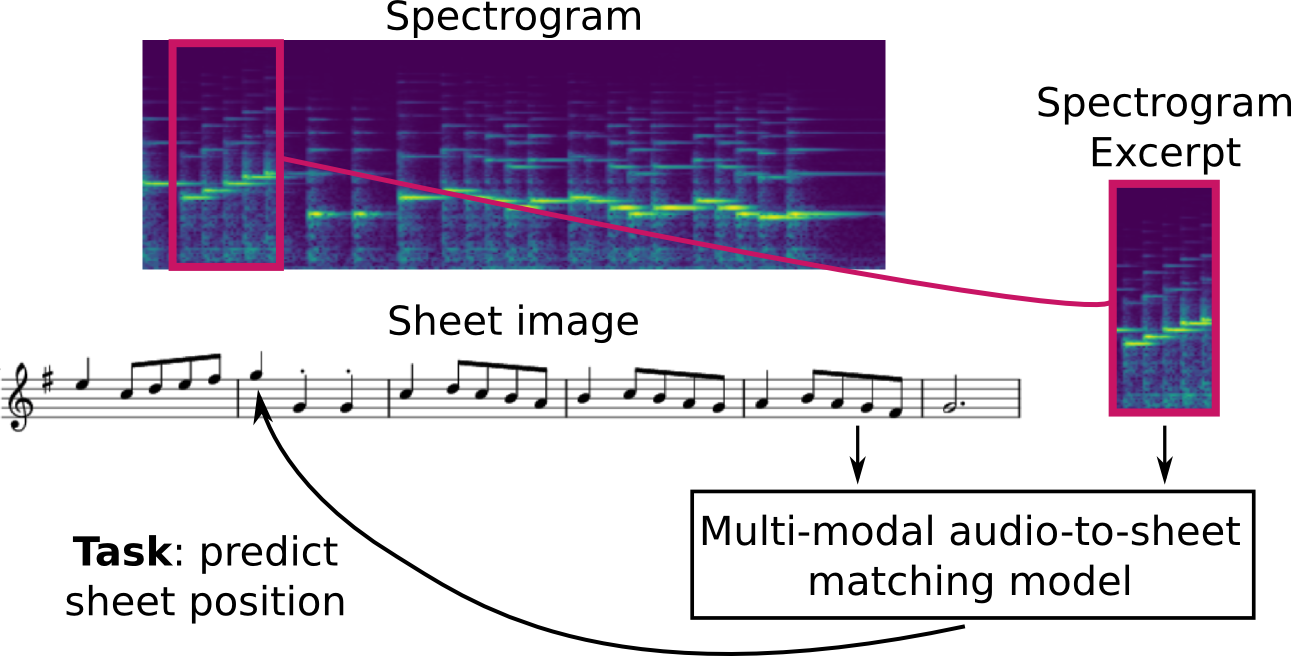}}}
\caption{\small Input data and audio-to-sheet matching task.}
\label{fig:comparison_stl10_32}
\end{figure}

A training/testing example is thus composed of two inputs:
Input 1 is an image $\mathbf{S}_i$ (in our case of size $40 \times 390$ pixels) showing one 
staff of sheet music.
Input 2 is an audio snippet
-- specifically, a spectrogram excerpt $\mathbf{E}_{i,j}$ ($40$ frames $\times$ $136$ frequency bins) --
cut from a recording of the piece, of fixed length ($1.2$ seconds).
The rightmost onset in spectrogram excerpt $\mathbf{E}_{i,j}$ is interpreted as the target note $j$
whose position we want to predict in staff image $\mathbf{S}_i$.
For the music used in our experiments (Section \ref{sec:experiments}) this context is a
bit less than one bar.
For each note $j$ (represented by its corresponding spectrogram excerpt $\mathbf{E}_{i,j}$)
we annotated its \emph{ground truth} sheet location $x_j$ in sheet image $\mathbf{S}_i$.
Coordinate $x_j$ is the distance of the note head (in pixels) from the left border of the image.
As we work with single staffs of sheet music
we only need the $x$-coordinate of the note at this point.
Figure \ref{fig:annotation_example} relates all components involved.

\textit{Summary and Task Description}: For training we present triples of (1) staff image $\mathbf{S}_i$, (2) spectrogram excerpt $\mathbf{E}_{i,j}$ and (3) ground truth pixel x-coordinate $x_j$ to our audio-to-sheet matching model.
At test time only the staff image and spectrogram excerpt are available and
the task of the model is to predict the estimated pixel location $\hat{x}_j$ in the image.
Figure \ref{fig:task_sketch} shows a sketch summarizing this task.

\subsection{Audio-Sheet Matching as Bucket Classification}
\label{subsec:bucket_clf}
We now propose a multi-modal convolutional neural network architecture
that learns to match unseen audio snippets (spectrogram excerpts) to their corresponding pixel location in the sheet image.

\subsubsection{Network Structure}
\begin{figure*}[ht!]
 \centerline{\includegraphics[width=1.5\columnwidth]{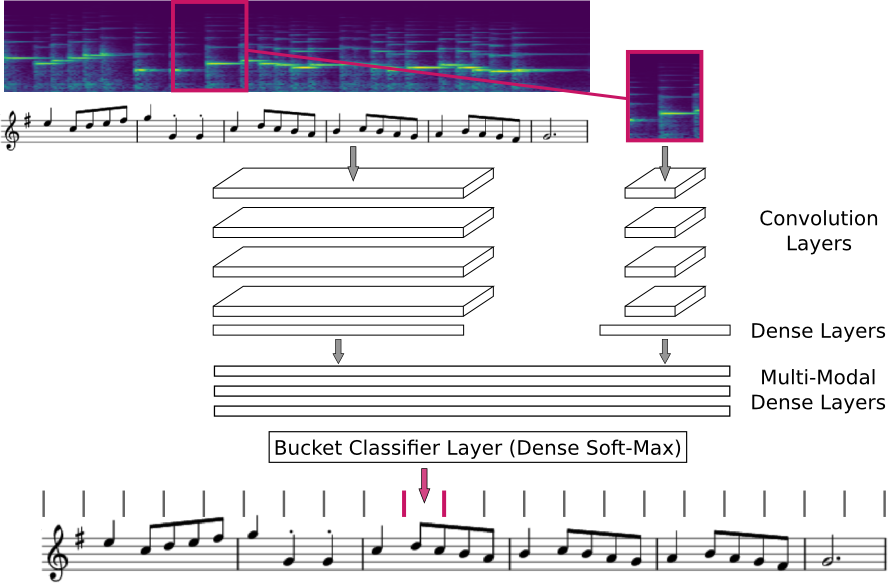}}
 \caption{\small Overview of multi-modal convolutional neural network for audio-to-sheet matching.
          The network takes a staff image and a spectrogram excerpt as input.
 		  Two specialized convolutional network parts, one for the sheet image
 		  and one for the audio input, are merged into one multi-modality network.
 		  The output part of the network predicts the region in the sheet image
 		  -- the classification bucket -- to which the audio snippet corresponds.
 		  % (See Section \ref{subsec:bucket_clf} for a detailed description.)
        }
\label{fig:model_architecture}
\end{figure*}

Figure \ref{fig:model_architecture} provides a general overview of the deep network
and the proposed solution to the matching problem.
As mentioned above, the model operates jointly on a staff image $\mathbf{S}_i$
and the audio (spectrogram) excerpt $\mathbf{E}_{i,j}$ related to a note $j$.
The rightmost onset in the spectrogram excerpt is the one related to target note $j$.
The multi-modal model consists of two specialized convolutional networks: one dealing
with the sheet image and one dealing with the audio (spectrogram) input.
In the subsequent layers we fuse the specialized sub-networks
by concatenation of the latent image- and audio representations
and additional processing by a sequence of dense layers.
For a detailed description of the individual layers
we refer to Table \ref{tab:model_architecture} in Section \ref{subsec:exp_setup}.
The output layer of the network and the corresponding localization principle
are explained in the following.

\subsubsection{Audio-to-Sheet Bucket Classification}
\begin{figure}[tb!]
 \centerline{\includegraphics[width=0.57\columnwidth]{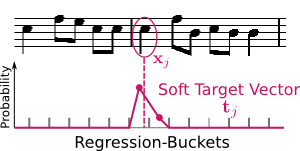}}
 \caption{\small Part of a staff of sheet music along with soft target vector $\mathbf{t}_j$ for target note $j$ surrounded with an ellipse.
 The two buckets closest to the note share the probability (indicated as dots) of containing the note.
 The short vertical lines highlight the bucket borders.}
 \label{fig:soft_targets}
\end{figure}

The objective for an unseen spectrogram excerpt and a corresponding staff of sheet music is
to predict the excerpt's location $x_j$ in the staff image.
For this purpose we start with horizontally quantizing the sheet image into $B$ non-overlapping buckets.
This discretisation step is indicated as the short vertical lines
in the staff image above the score in Figure \ref{fig:model_architecture}.
In a second step we create for each note $j$ in the train set a target vector $\mathbf{t}_j=\{t_{j,b}\}$
where each vector element $t_{j,b}$ holds the probability that bucket $b$ covers the current target note $j$.
In particular, we use soft targets, meaning that the probability for one note
is shared between the two buckets closest to the note's true pixel location $x_j$.
We linearly interpolate the shared probabilities based on the two pixel distances
(normalized to sum up to one) of the note's location $x_j$ to the respective (closest) bucket centers.
Bucket centers are denoted by $c_b$ in the following where subscript $b$ is the index of the respective bucket.
Figure \ref{fig:soft_targets} shows an example sketch of the components described above.
Based on the soft target vectors we design the output layer of our audio-to-sheet matching network
as a $B$-way soft-max with activations defined as:
\begin{equation}
\phi(y_{j,b}) = \frac{e^{y_{j,b}}}{\sum_{k=1}^B e^{y_{j,k}}}
\label{eq:softmax_output}
\end{equation}
$\phi(y_{j,b})$ is the soft-max activation of the output neuron representing bucket $b$
and hence also representing the region in the sheet image covered by this bucket.
By applying the soft-max activation the network output gets normalized to range $(0,1)$
and further sums up to $1.0$ over all $B$ output neurons.
The network output can now also be interpreted as a vector of probabilities $\mathbf{p}_{j}=\{\phi(y_{j,b})\}$
and shares the same value range and properties as the soft target vectors.

In training, we optimize the network parameters $\Theta$
by minimizing the Categorical Cross Entropy (CCE) loss $l_j$
between target vectors $\mathbf{t}_j$ and network output $\mathbf{p}_{j}$:
\begin{equation}
l_j(\Theta) = - \sum_{k=1}^{B} t_{j,k} \, log(p_{j,k})
\end{equation}
The CCE loss function becomes minimal when the network output $\mathbf{p}_{j}$
exactly matches the respective soft target vector $\mathbf{t}_j$.
In Section \ref{subsec:exp_setup} we provide further information on the exact optimization strategy used.\footnote{
\textit{For the  sake of completeness}:
In our initial experiments we started to predict the sheet location of audio snippets
by minimizing the Mean-Squared-Error (MSE) between the predicted and the true pixel coordinate (MSE regression).
However, we observed that training these networks is much harder
and further performs worse than the bucket classification approach proposed in this paper.}

\subsection{Sheet Location Prediction}
\label{subsec:location_pred}
Once the model is trained, we use it at test time to predict the expected location $\hat{x}_j$
of an audio snippet with target note $j$ in a corresponding image of sheet music.
The output of the network is a vector $\mathbf{p}_{j} = \{p_{j,b}\}$
holding the probabilities that the given test snippet $j$ matches with bucket $b$ in the sheet image.
Having these probabilities we consider two different types of predictions:
(1) We compute the center $c^*_{b}$ of bucket $b^*=\text{argmax}_{b} \, p_{j,b}$
holding the highest overall matching probability.
(2) For the second case we take, in addition to $b^*$, the two neighbouring buckets $b^*-1$ and $b^*+1$ into account and
compute a (linearly) probability weighted position prediction in the sheet image as
\begin{equation}
\hat{x}_j = \sum_{k \in \{b^*-1, b^*, b^*+1\}} w_{k}c_{k}
\end{equation}
where weight vector $\mathbf{w}$ contains the probabilities $\{p_{j,b^*-1}, p_{j,b^*}, p_{j,b^*+1}\}$
normalized to sum up to one and $c_{k}$ are the center coordinates of the respective buckets.

\section{Experimental Evaluation}
\label{sec:experiments}
%!TEX root = ISMIR2016template.tex
This section evaluates our audio-to-sheet matching model on a
publicly available dataset. We describe the experimental setup, including the
data and evaluation measures, the particular network architecture as well as the
optimization strategy, and provide quantitative results.

\begin{table*}[ht]
\begin{center}
\begin{tabular}{c|c}
\hline
Sheet-Image $40 \times 390$ & Spectrogram $136 \times 40$ \\
\hline
$5\times5$ Conv(pad-2, stride-1-2)-$64$-BN-ReLu 		& $3\times3$ Conv(pad-1)-$64$-BN-ReLu \\
$3\times3$ Conv(pad-1)-$64$-BN-ReLu					& $3\times3$ Conv(pad-1)-$64$-BN-ReLu \\
$2\times2$ Max-Pooling + Drop-Out($0.15$)			& $2\times2$ Max-Pooling + Drop-Out($0.15$) \\
$3\times3$ Conv(pad-1)-$128$-BN-ReLu					& $3\times3$ Conv(pad-1)-$96$-BN-ReLu \\
$3\times3$ Conv(pad-1)-$128$-BN-ReLu					& $2\times2$ Max-Pooling + Drop-Out($0.15$) \\
$2\times2$ Max-Pooling + Drop-Out($0.15$)			& $3\times3$ Conv(pad-1)-$96$-BN-ReLu \\
													& $2\times2$ Max-Pooling + Drop-Out($0.15$) \\
Dense-$1024$-BN-ReLu + Drop-Out($0.3$)				& Dense-$1024$-BN-ReLu + Drop-Out($0.3$) \\
\hline
\multicolumn{2}{c}{Concatenation-Layer-$2048$} \\
\multicolumn{2}{c}{Dense-$1024$-BN-ReLu + Drop-Out($0.3$)} \\
\multicolumn{2}{c}{Dense-$1024$-BN-ReLu + Drop-Out($0.3$)} \\
\hline
\multicolumn{2}{c}{$B$-way Soft-Max Layer} \\
\end{tabular}
\end{center}
\caption{\small Architecture of Multi-Modal Audio-to-Sheet Matching Model:
BN: Batch Normalization, ReLu: Rectified Linear Activation Function,
CCE: Categorical Cross Entropy, Mini-batch size: 100}
\label{tab:model_architecture}
\end{table*}

% \subsection{Experimental Setup}
% \label{subsec:exp_setup}
% In this section we introduce the data and explain how we prepare it (midi-to-sheet rendering, midi synthesizing, note annotation, ...)
% to be suitable for the evaluation of our approach.
% We further introduce the evaluation measures used to quantify the matching performance of our model
% and provide details on the particular network architecture as well as the optimization strategy used.

\subsection{Experiment Description}
The aim of this paper is to show that it is feasible to learn correspondences
between audio (spectrograms) and images of sheet music in  an
\emph{end-to-end} neural network fashion, meaning that
an algorithm learns the entire task purely from data, so that
no hand crafted feature engineering is required.
We try to keep the experimental setup simple
and consider one staff of sheet music per train/test sample (this is exactly the setup drafted in Figure \ref{fig:model_architecture}).
To be perfectly clear, the task at hand is the following:
For a given audio snippet, find its x-coordinate pixel position in a corresponding staff of sheet music.
We further restrict the audio to monophonic music containing half, quarter and eighth notes but allow
variations such as dotted notes, notes tied across bar lines as well as accidental signs.

\subsection{Data}
For the evaluation of our approach we consider the Nottingham\footnote{\url{www-etud.iro.umontreal.ca/~boulanni/icml2012}} data set
which was used, e.g., for piano transcription in \cite{Boulanger_2012_ModSequ}.
It is a collection of midi files already split into train, validation and test tracks.
To be suitable for audio-to-sheet matching we prepare the data set (midi files) as follows:
\begin{enumerate}
\item We select the first track of the midi files (right hand, piano) and render it as sheet music using Lilypond.\footnote{\url{http://www.lilypond.org/}}
\item We annotate the sheet coordinate $x_j$ of each note.
%\footnote{We implemented a semi-automatic script that annotates both x- and y- pixel-coordinate to have it available for further research.}
\item We synthesize the midi-tracks to \textit{flac}-audio using Fluidsynth\footnote{\url{http://www.fluidsynth.org/}} and a \emph{Steinway} piano sound font.
\item We extract the audio timestamps of all note onsets.
\end{enumerate}
%
% We now have all the components required to evaluate our approach.
As a last preprocessing step we compute \emph{log-spectrograms} of the synthesized
flac files \cite{Boeck_2016_madmom},
% using the \emph{madmom} audio signal processing library,% \footnote{\url{https://github.com/CPJKU/madmom}}
with an audio sample rate of $22.05$kHz, FFT window size of $2048$ samples,
and computation rate of $31.25$ frames per second.
For dimensionality reduction we apply a normalized $24$-band logarithmic filterbank
allowing only frequencies from $80$Hz to $8$kHz. This results in $136$ frequency bins.

We already showed a spectrogram-to-sheet annotation example in Figure \ref{fig:annotation_example}.
% used for training the multi-modal matching model.
In our experiment we use spectrogram excerpts covering $1.2$ seconds of audio (40 frames).
This context is kept the same for training and testing.
Again, annotations are aligned in a way so that the rightmost onset in a spectrogram excerpt
corresponds to the pixel position of target note $j$ in the sheet image.
In addition, the spectrogram is shifted 5 frames to the right to also contain some information
on the current target note's onset and pitch.
We chose this annotation variant with the rightmost onset
as it allows for an online application of our audio-to-sheet model
(as would be required, e.g., in a score following task).

\subsection{Evaluation Measures}

To evaluate our approach we consider, for each test note $j$, the following ground truth and prediction data:
(1) The true position $x_j$ as well as the corresponding target bucket $b_j$ (see Figure \ref{fig:soft_targets}).
(2) The estimated sheet location $\hat{x}_j$ and the most likely target bucket $b^*$ predicted by the model.
Given this data we compute two types of evaluation measures.

The first -- the \emph{top-k bucket hit rate} --
quantifies the ratio of notes that are classified into the correct bucket allowing a tolerance of $k-1$ buckets.
For example, the \emph{top-1 bucket hit rate} counts only those notes
where the predicted bucket $b^*$ matches exactly the note's target bucket $b_j$.
The \emph{top-2 bucket hit rate} allows for a tolerance of one bucket and so on.
The second measure -- the \emph{normalized pixel distance} -- captures the actual distance of a predicted sheet location $\hat{x}_j$
to its corresponding true position $x_j$.
To allow for an evaluation independent of the image resolution used in our experiments
we normalize the pixel errors by dividing them
by the width of the sheet image as $(\hat{x}_j - x_j) / width(\mathbf{S}_i)$.
This results in distance errors living in range $(-1, 1)$.

We would like to emphasise that the quantitative evaluations based on the measures introduced above are
performed only at time steps where a note onset is present.
At those points in time an explicit correspondence between spectrogram (onset) and sheet image (note head) is established.
However, in Section \ref{sec:discussion} we show that a time-continuous prediction
is also feasible with our model and onset detection is not required at run time.

\subsection{Model Architecture and Optimization}
\label{subsec:exp_setup}
Table \ref{tab:model_architecture} gives details on the model architecture used for our experiments.
As shown in Figure \ref{fig:model_architecture},
the model is structured into two disjoint convolutional networks
where one considers the sheet image and one the spectrogram (audio) input.
The convolutional parts of our model are inspired by the VGG model
built from sequences of small convolution kernels (e.g. $3 \times 3$) and max-pooling layers.
The central part of the model consists of a concatenation layer
bringing the image and spectrogram sub-networks together.
After two dense layers with 1024 units each we add a $B$-way soft-max output layer.
Each of the $B$ soft-max output neurons corresponds to one of the disjoint buckets
which in turn represent quantised sheet image positions.
In our experiments we use a fixed number of $40$ buckets selected as follows:
We measure the minimum distance between two subsequent notes -- in our sheet renderings --
and select the number of buckets such that each bucket contains at most one note.
It is of course possible that no note is present in a bucket --
e.g., for the buckets covering the clef at the beginning of a staff.
As activations function for the inner layers we use rectified linear units \cite{Glorot_2011_ReLU}
and apply batch normalization \cite{Loffe_2015_BatchNorm} after each layer as it helps training and convergence.

\begin{figure}[t]
 \centerline{\includegraphics[width=\columnwidth]{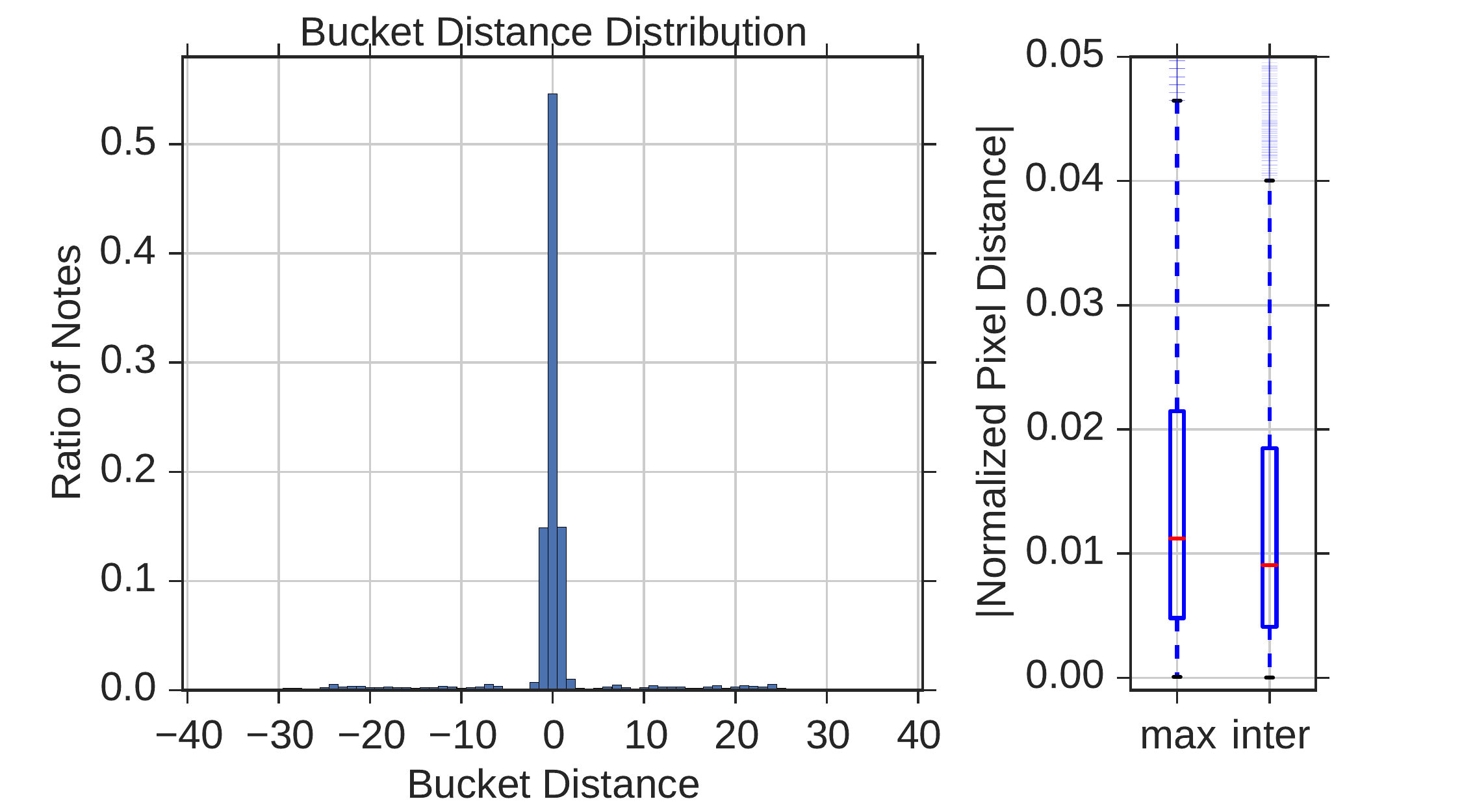}}
 \caption{\small Summary of matching results on test set.
 		  \textit{Left}: Histogram of bucket distances between predicted and true buckets.
 		  \textit{Right}: Box-plots of absolute \emph{normalized pixel distances} between predicted and true image position.
 		  The box-plot is shown for both location prediction methods described in Section \ref{subsec:location_pred} (maximum, interpolated).}
 \label{fig:bucket_hist}
\end{figure}

Given this architecture and data we optimize the parameters of the model using
mini-batch stochastic gradient descent with Nesterov style momentum.
We set the batch size to $100$ and fix the momentum at $0.9$ for all epochs.
The initial learn-rate is set to $0.1$ and divided by 10 every 10 epochs.
We additionally apply a weight decay of $0.0001$ to all trainable parameters of the model.

\subsection{Experimental Results}
\label{subsec:exp_results}

Figure \ref{fig:bucket_hist} shows a histogram of the signed bucket distances
between predicted and true buckets.
% of a note $j$'s
% predicted bucket $b^*$ to its true bucket $b_j$.
The plot shows that more than $54\%$ of all unseen test notes are matched exactly with the corresponding bucket.
When we allow for a tolerance of $\pm 1$ bucket our model is able to assign over $84\%$ of the test notes correctly.
We can further observe that the prediction errors are equally distributed in both directions
-- meaning too early and too late in terms of audio.
The results are also reported in numbers  in Table \ref{tab:bucket_hit_rates},
as the top-k bucket hit rates for train, validation and test set.

\begin{table}[t]
 \begin{center}
 \begin{tabular}{l|lll}
  & \textbf{Train} & \textbf{Valid} & \textbf{Test} \\
  \hline
  \hline
  Top-1-Bucket-Hit-Rate 		& 79.28\%	& 51.63\%	& 54.64\% \\
  Top-2-Bucket-Hit-Rate 		& 94.52\%	& 82.55\%	& 84.36\% \\
  \hline
  mean($|NPD_{max}|$) 		& 0.0316		& 0.0684		& 0.0647 \\
  mean($|NPD_{int}|$) 		& 0.0285		& 0.0670		& 0.0633 \\
  median($|NPD_{max}|$)		& 0.0067		& 0.0119		& 0.0112 \\
  median($|NPD_{int}|$) 		& 0.0033		& 0.0098		& 0.0091 \\
  \hline
  $|NPD_{max}|< w_{b}$ 		& 93.87\%		& 76.31\%	& 79.01\% \\
  $|NPD_{int}|< w_{b}$ 		& 94.21\%		& 78.37\%	& 81.18\% \\
 \end{tabular}
\end{center}
 \caption{\small \emph{Top-k bucket hit rates} and \emph{normalized pixel distances} (NPD) as described in Section \ref{subsec:exp_setup} for train, validation and test set.
          We report mean and median of the absolute NPDs for both interpolated (int) and maximum (max) probability bucket prediction.
          The last two rows report the percentage of predictions not further away from the true pixel location than the width $w_b$ of one bucket.}
 \label{tab:bucket_hit_rates}
\end{table}

The box plots in the right part of Figure \ref{fig:bucket_hist} summarize the absolute \emph{normalized pixel distances (NPD)} between predicted and true locations.
We see that the probability-weighted position interpolation (Section \ref{subsec:location_pred})
helps improve the localization performance of the model.
Table \ref{tab:bucket_hit_rates} again puts the results in numbers,
as means and medians of the absolute NPD values.
Finally, Fig.~\ref{tab:bucket_hit_rates} (bottom) reports the ratio of predictions
with a pixel distance smaller than the width of a single bucket.
% In addition, we investigate the actual predictions
% of the proposed audio-to-sheet matching model in detail in the next section.
%

\section{Discussion and Real Music}
\label{sec:discussion}
%!TEX root = ISMIR2016template.tex
This section provides a representative prediction example of our model
and uses it to discuss the proposed approach.
In the second part we then show a first step towards matching \emph{real}
(though still very simple) music to its corresponding sheet.
By \emph{real music} we mean audio that is not just synthesized midi,
but played by a human on a piano and recorded via microphone.
% We also provide a link to a video showing our multi-modal audio-to-sheet
% matching network in action.

\begin{figure*}[!t]
 \centerline{\includegraphics[width=0.98\textwidth]{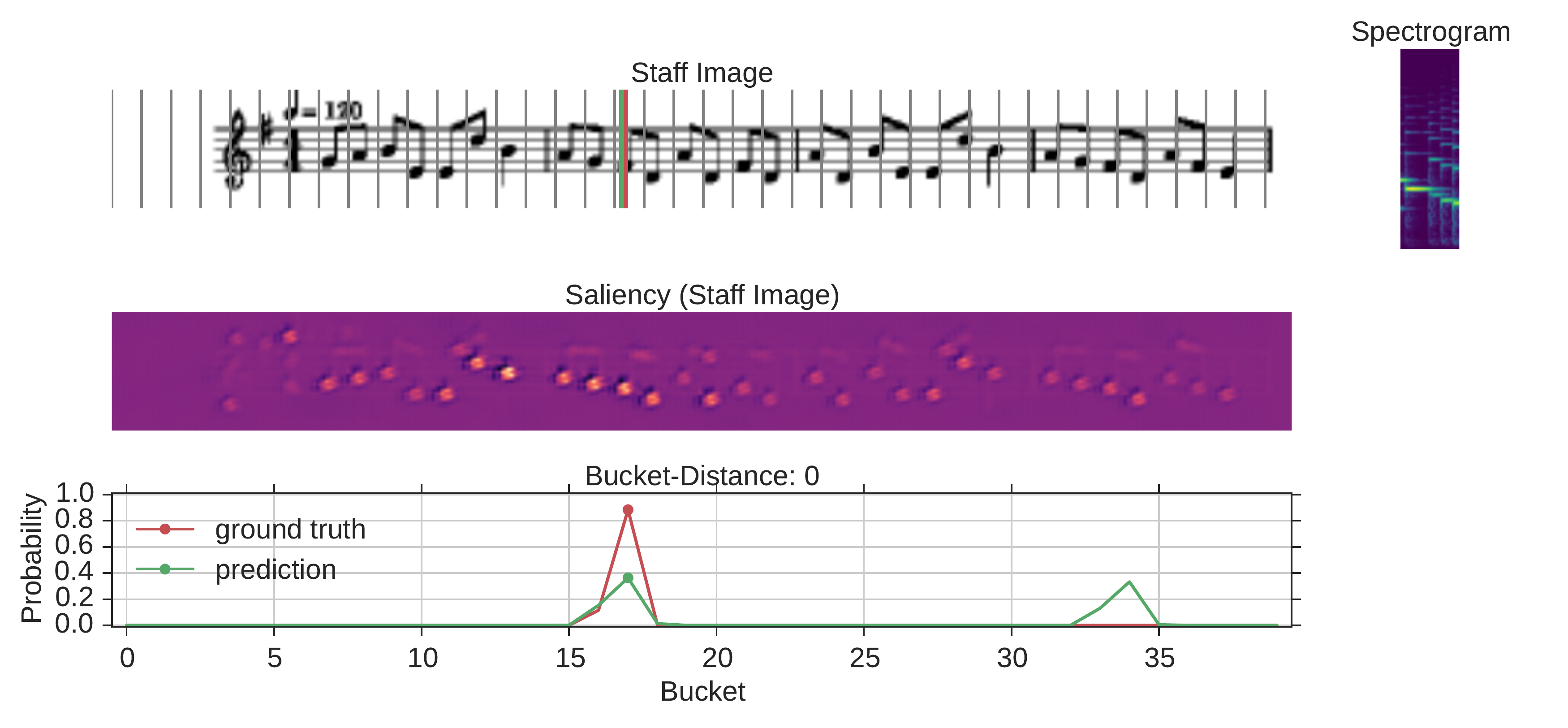}}
 \caption{\small Example prediction of the proposed model.
          The top row shows the input staff image $\mathbf{S}_i$ along with the bucket borders as thin gray lines,
          and the given query audio (spectrogram) snippet $\mathbf{E}_{i,j}$.
          The plot in the middle visualizes the salience map (representing the attention of the neural network) computed on the input image.
          Note that the network's attention is actually drawn to the individual note heads.
          The bottom row compares the ground truth bucket probabilities with the probabilities predicted by the network.
          In addition, we also highlight the corresponding true and predicted pixel locations in the staff image in the top row.}
 \label{fig:discussion_example}
\end{figure*}

\subsection{Prediction Example and Discussion}
Figure \ref{fig:discussion_example} shows the image of one staff of sheet music
along with the predicted as well as the ground truth pixel location for a snippet of audio.
The network correctly matches the spectrogram with the corresponding pixel location in the sheet image.
However, we observe a second peak in the bucket prediction probability vector.
A closer look shows that this is entirely reasonable, as the music is quite repetitive
and the current target situation actually appears twice in the score.
The ability of predicting probabilities for multiple positions
is a desirable and important property, as repetitive structures are immanent to music.
The resulting prediction ambiguities
can be addressed by exploiting the temporal relations between the notes
in a piece by methods such as dynamic time warping or probabilistic models.
In fact, we plan to combine the probabilistic output of our matching model with existing score following methods, as for example \cite{Arzt_2008_Page}.
In Section \ref{sec:methods} we mentioned that training a sheet location prediction
with MSE-regression is difficult to optimize.
Besides this technical drawback it would not be straightforward
to predict a variable number of locations with an MSE-model,
as the number of network outputs has to be fixed when designing the model.

In addition to the network inputs and prediction Fig.~\ref{fig:discussion_example}
also shows a \emph{saliency map} \cite{Springenberg_2014_AllConv}
computed on the input sheet image with respect to the
network output.\footnote{The implementation is adopted from an example by Jan Schl{\"u}ter in the recipes section of the deep learning framework \emph{Lasagne} 
\cite{Dieleman_2015_Lasagne}.}
The saliency can be interpreted as the input regions to which most of the net's attention is drawn.
In other words, it highlights the regions that contribute most to the current output produced by the model.
A nice insight of this visualization is
that the network actually focuses and recognizes the heads of the individual notes.
In addition it also directs some attention to the style of stems,
which is necessary to distinguish for example between quarter and eighth notes.

The optimization on soft target vectors is also
reflected in the predicted bucket probabilities.
In particular the neighbours of the bucket with maximum activation are also active
even though there is no explicit neighbourhood relation encoded in the soft-max output layer.
This helps the interpolation of the true position in the image (see Fig.~\ref{fig:bucket_hist}).

\subsection{First Steps with Real Music}
As a final point, we report on first attempts at working with ``real'' music.
For this purpose one of the authors played the right hand part of a simple piece
(Minuet in G Major by Johann Sebastian Bach, BWV Anhang 114) -- which, of course,
was not part of the training data --
on a \emph{Yamaha AvantGrand N2} hybrid piano and recorded it using a single microphone.
In this application scenario we predict the corresponding sheet locations not only at times of onsets
but for a continuous audio stream (subsequent spectrogram excerpts).
This can be seen as a simple version of online score following in sheet music,
without taking into account the temporal relations of the predictions.
We offer the reader a video\footnote{\url{https://www.dropbox.com/s/0nz540i1178hjp3/Bach_Minuet_G_Major_net4b.mp4?dl=0}}
that shows our model
following the first three staff lines of this simple piece.\footnote{
Note: our model operates on single staffs of sheet music and
requires a certain context of spectrogram frames for prediction (in our case 40 frames).
For this reason it cannot provide a localization for the first couple of notes in the beginning
of each staff at the current stage.
In the video one can observe that prediction only starts when the spectrogram in the top right corner
has grown to the desired size of 40 frames.
We kept this behaviour for now as we see our work as a proof of concept.
The issue can be easily addressed by concatenating the images of subsequent staffs in horizontal direction.
In this way we will get a ``continuous stream of sheet music'' analogous to a spectrogram for audio.}
The ratio of predicted notes having a pixel-distance smaller than the bucket width (compare Section \ref{subsec:exp_results})
is $71.72$\% for this real recording.
This corresponds to a average normalized-pixel-distance of $0.0402$.

\section{Conclusion}
\label{sec:conclusion}
%!TEX root = ISMIR2016template.tex
In this paper we presented a multi-modal convolutional neural network which is
able to match short snippets of audio with their corresponding position in
the respective image of sheet music, without the need of any symbolic
representation of the score. First evaluations on simple piano music suggest
that this is a very promising new approach that deserves to be explored further.

As this is a proof of concept paper, naturally our method still has some severe
limitations. So far our approach can only deal with monophonic music, notated on
a single staff, and with performances that are roughly played in the same tempo
as was set in our training examples.

In the future we will explore options to lift these limitations one by one, with
the ultimate goal of making this approach applicable to virtually any kind of
complex sheet music. In addition, we will try to combine this approach with a
score following algorithm. Our vision here is to build a score following system
that is capable of dealing with any kind of classical sheet music, out of the box,
with no need for data preparation.

\section{Acknowledgements}
This work is supported by the Austrian Ministries BMVIT and BMWFW,
and the Province of Upper Austria via the COMET Center SCCH,
and by the European Research Council (ERC Grant Agreement
670035, project CON ESPRESSIONE).
The Tesla K40 used for this research was donated by the NVIDIA
corporation.

% For bibtex users:
\bibliography{ISMIRtemplate}

% For non bibtex users:
%\begin{thebibliography}{citations}
%
%\bibitem {Author:00}
%E. Author.
%``The Title of the Conference Paper,''
%{\it Proceedings of the International Symposium
%on Music Information Retrieval}, pp.~000--111, 2000.
%
%\bibitem{Someone:10}
%A. Someone, B. Someone, and C. Someone.
%``The Title of the Journal Paper,''
%{\it Journal of New Music Research},
%Vol.~A, No.~B, pp.~111--222, 2010.
%
%\bibitem{Someone:04} X. Someone and Y. Someone. {\it Title of the Book},
%    Editorial Acme, Porto, 2012.
%
%\end{thebibliography}

\end{document}